%% file: rldm.tex
\documentclass[11pt]{article} 
\usepackage{rldmsubmit,palatino}
\usepackage{graphicx}

\usepackage[utf8]{inputenc}
\usepackage{amsmath, amssymb, latexsym}
\usepackage{sidecap}
\usepackage{gensymb}

\usepackage{tikz}
\usetikzlibrary{decorations.pathreplacing}
\usepackage[outline]{contour}
\contourlength{0.9pt}
\definecolor{lightgray}{gray}{0.95}

\usepackage{caption}
\usepackage{booktabs}

\usepackage[backend=biber, url=false, style=alphabetic]{biblatex}
\bibliography{bibliography}

\title{Deep Reinforcement Learning with Model Learning\\and Monte Carlo Tree Search in Minecraft}

\author{Stephan Alaniz\\
Department of Electrical Engineering and Computer Science\\
Technische Universit\"at Berlin\\
Berlin, Germany\\}

\begin{document}

\maketitle

\begin{abstract}
Deep reinforcement learning has been successfully applied to several visual-input tasks using model-free methods.  In this paper, we propose
a model-based approach that combines learning a DNN-based transition model with
Monte Carlo tree search to solve a block-placing task in Minecraft.  Our
learned transition model predicts the next frame and the rewards one step ahead
given the last four frames of the agent's first-person-view image and the
current action. Then a Monte Carlo tree search algorithm uses this model to
plan the best sequence of actions for the agent to perform.  On the proposed
task in Minecraft, our model-based approach reaches the performance comparable
to the Deep Q-Network's, but learns faster and, thus, is more training sample efficient.

\end{abstract}

\keywords{Reinforcement Learning, Model-Based Reinforcement Learning, Deep Learning, Model Learning, Monte Carlo Tree Search}

\acknowledgements{}

I would like to express my sincere gratitude to my supervisor Dr. Stefan Uhlich for his continuous support, patience, and immense knowledge that helped me a lot during this study. My thanks and appreciation also go to my colleague Anna Konobelkina for insightful comments on the paper as well as to Sony Europe Limited for providing the resources for this project.

\startmain 

\section{Introduction}
In deep reinforcement learning, visual-input tasks (i.e., tasks where the observation from the environment comes in the form of videos or pictures) are oftentimes used to evaluate algorithms: Minecraft or various Atari games appear quite challenging for agents to solve~\cite{oh2016minecraft},~\cite{mnih2015dqn}. When applied to these tasks, model-free reinforcement learning shows noticeably good results: e.g., a Deep Q-Network (DQN) agent approaches human-level game-playing performance~\cite{mnih2015dqn} or the Asynchronous Advantage Actor-Critic algorithm outperforms the known methods in half the training time~\cite{mnih2016a3c}. These achievements, however, do not cancel the fact that generally, model-free methods are considered “statistically less efficient” in comparison to model-based ones: model-free approaches do not employ the information about the environment directly whereas model-based solutions do~\cite{dayan2008}.

Working with a known environment model has its benefits: changes of the environment state can be foreseen, therefore, planning the future becomes less complicated. At the same time, developing algorithms with no known environment model available at the start is more demanding yet promising: less training data is required than for model-free approaches and agents can utilize planning algorithms. The research has been progressing in this direction: e.g., a model-based agent surpassed the DQN's results by using the Atari games’ true state for modelling~\cite{xiaoxiao2014mcts}, and constructing transition models with video-frames prediction was proposed~\cite{oh2015frame}.

These ideas paved the way for the following question: is it feasible to apply planning algorithms on a learned model of the environment that is only partially observable, such as in a Minecraft building task? To investigate this question, we developed a method that not only predicts future frames of a visual task but also calculates the possible rewards for the agent’s actions. Our model-based approach merges model learning through deep-neural-network training with Monte Carlo tree search, and demonstrates results competitive with those of DQN’s, when tested on a block-placing task in Minecraft.

\input{mcfig}
\section{Block-Placing Task in Minecraft}

To evaluate the performance of the suggested approach as well as to compare it with model-free methods, namely, DQN, a block-placing task was designed: it makes use of the Malmo framework and is built inside the Minecraft game world~\cite{malmo2016}.

At the beginning of the game, the agent is positioned to the wall of the playing ``room''. There is a 5\(\times\)5 playing field in the center of this room. The field is white with each tile having a 0.1 probability to become colored at the start. Colored tiles indicate the location for the agent to place a block.
The goal of the game is to cover all the colored tiles with blocks in 30 actions maximum. Five position-changing actions are allowed: moving forward by one tile, turning left or right by 90\degree, and moving sideways to the left or right by one tile. When the agent focuses on a tile, it can place the block with the 6th action.
For each action, the agent receives a feedback: every correctly placed block
brings a +1 reward, an erroneously put block causes a -1 punishment, and any
action costs the agent a -0.04 penalty (this reward signal is introduced to
stimulate the agent to solve the task with the minimum time required).  
To evaluate the environment, the agent is provided with a pre-processed
(grayscaled and downsampled to 64\(\times\)64) first-person-view picture of the
current state. The task is deterministic and discrete in its action and state space. The challenge lies in the partial observability of the environment with already
placed blocks further obscuring the agent’s view. It is equally important to place blocks systematically to not obstruct the agent’s pathway. 
An example of the task is depicted in Fig.~\ref{fig:minecraft} (left) and a short demonstration is available at https://youtu.be/AQlBaq34DpA.\@

\input{dnn}
\section{Model Learning}\label{sec:dnn}
To learn the transition model, a deep convolutional neural network is used.
The network takes the last four frames \(s_{t-3}, \dots, s_{t}\), and an action
\(a_t\) as an input, and predicts the following frame \(\widehat{s}_{t+1}\).
Additionally, it predicts the rewards for all the transitions following the
predicted frame \(\widehat{s}_{t+1}\), one for each action \(a_{t+1}\). 
Predicting the rewards one step ahead makes the application of search-based
algorithms more efficient as no additional simulation is required to explore
rewards from transitions to neighboring states. 
This method, however, fails to predict the reward following the very first
state. To address this issue, a ``noop'' action, predicting the reward of the
current state, is introduced.

The network takes four 64\(\times\)64-sized input
frames and uses four convolutional layers, each followed
by a rectifier linear unit (ReLU), to encode the input information into a
vector of the size of 4096. This vector is concatenated with the one-hot
encoded action input, where the ``noop'' action is represented by a vector of
all zeros, and then linearly transformed with a fully connected layer of the size of 4096, again followed by ReLU\@. 
The resulting embedded vector is used for both, the reward prediction and the
frame prediction, in the last part of the network.  For the frame prediction,
four deconvolutional layers are used
with ReLUs in between and a Sigmoid at the end. The dimensions of these layers
are equivalent to the convolutional layers in the first part in reversed
order. The reward prediction is done by applying two fully connected linear layers
of the sizes of 2048 and 6, respectively. ReLU is used in between the two layers, but not
at the final output where we predict the rewards. The architecture of the
neural network is illustrated in Fig.~\ref{fig:dnnarch}.

Training of the network is done with the help of experience
replay~\cite{mnih2015dqn} to re-use and de-correlate the training data. Mini-batches
of the size of 32 are sampled for training and RMSProp~\cite{Tieleman2012} is used to
update the weights. Both, the frame prediction and the reward prediction, are trained with a mean squared error (MSE) loss.  On each mini-batch update, the gradients of the frame prediction loss are backpropagated completely through the network while the gradients of the reward prediction loss are backpropagated only two layers until the embedded vector shared with the frame prediction is encountered.  This procedure ensures that the network uses
its full capacity to improve the prediction of the next state and the reward
prediction is independently trained on the embedded feature vector from the
last shared intermediate layer. Due to the network's structure, the shared
layer may only contain the necessary information to construct the prediction of the next frame and no further past information. For the block-placing task
, one frame suffices to predict the reward. For different tasks, using the previous layer with the action input is worth considering (cf.
Fig.~\ref{fig:dnnarch}).

\section{Monte Carlo Tree Search}
Finding actions with maximum future reward is done with the help of a UCT-based
strategy, Monte Carlo tree search (MCTS)~\cite{coulom2006mcts, kocsis2006uct},
and the learned model described in Sec.~\ref{sec:dnn}. The input for the next step is
based on the frame prediction of the model. This procedure can be repeated several times
to roll out the model to future states. One tree-search trajectory is rolled
out until a maximum depth is reached.  During one rollout, all the  rewards along
the trajectory are predicted as well as the rewards for neighboring states due
to the network's structure. As the outputs of the neural network are
deterministic, each state is evaluated only once but can still be visited
multiple times since many paths go through the same states. The decision about
which action to follow for each state during a tree-search trajectory is made
based on the UCT-measure \(UCT_s = v_s + k \sqrt{(\ln n_p) / n_s}\).
The action with the maximum UCT value is chosen greedily, where \(v_s\)
is the maximum discounted future reward of the state \(s\) so far,
\(n_s\) is the number of visits of state \(s\), and \(n_p\) is the number of
visits of its parent. The hyperparameter \(k\) controls the trade-off of
exploration and exploitation, where a higher \(k\) translates to more
exploration. If a state has not been visited yet, it will be preferred over
the already visited ones. In case several states under consideration have
not been visited yet, the one with the highest immediate reward is chosen. The
reward is given by the neural network model that predicts rewards one step
ahead. Whenever a path of higher maximum value is encountered during a search
trajectory, the path's value is propagated to the tree's root node updating
every node with the new maximum reward value.

During the evaluation of the task in Minecraft, the agent performs MCTS
for each action decision it has to make. The agent is given a fixed number of
trajectories to roll out and decides for the action of maximum future
discounted reward to take as the next action. Subsequently, the agent receives
a new ground-truth input frame from the environment for the last step and
updates the root of the search tree with the new state information. The next
step is again chosen by applying MCTS beginning from the new state. Instead of
starting from scratch, the UCT value is calculated with the maximum future reward value of the previous turn. This way, the tree-search results are
carried over to following steps and trajectories with maximum future
reward are updated first with the new input.

\section{Experiments and Results}
\begin{figure}
    \centering
    \includegraphics[width=0.8\textwidth]{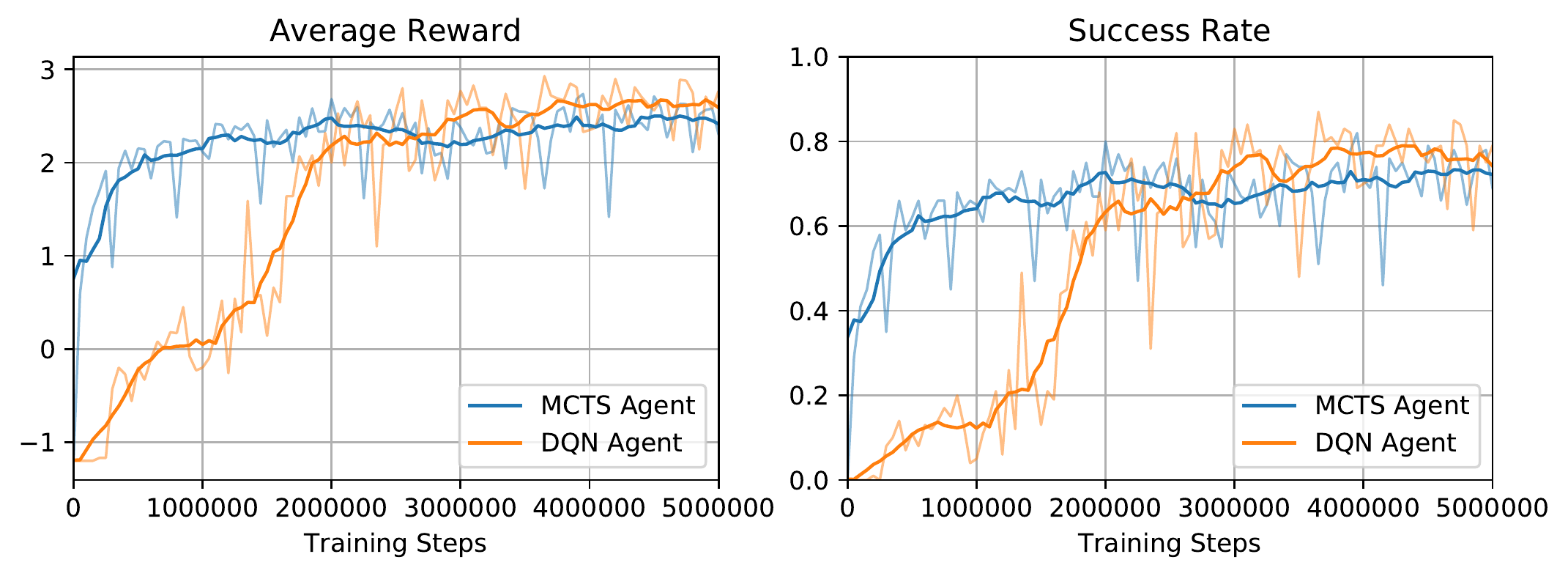}
    \vspace{-2.5mm}
    \captionsetup{format=hang}
    \caption{MCTS Agent vs. DQN Agent: Average Reward and Success Rate on 100 Block-Placing Tasks in Minecraft. \\
    	\small One training step = one mini-batch update; raw values (transparent lines, evaluated every 50K steps)\\smoothed with moving average (solid lines, 200K steps to each side).}\label{fig:mctsdqn}
\end{figure}

The MCTS
agent uses a maximum depth of ten for its rollouts and explores 100 trajectories
before deciding for the next action. Over the course of solving a single task, the
MCTS agent explores 3000 trajectories at most since the task is limited to 30
actions. Trajectories, however, may be repeated with new input frames arriving
after every action decision. The hyperparameter \(k\) of UCT is set to eight.
Compared to other domains, this value is rather large. The reason behind this
decision is as follows: our exploitation measure (maximum future reward) is not
limited in range, and the number of visits remain low since each node is
evaluated only once. We use the original DQN
architecture~\cite{mnih2015dqn} to train a DQN agent for the
block-placing task. Restricting the input to four frames for our method grounds
in the same number of frames used in the DQN's structure to provide both approaches with equal input information.
In Fig.~\ref{fig:mctsdqn}, the results for task-solving success rate and the average reward are presented.
For both agents, MCTS and DQN, the two performance measures were evaluated
over the course of training steps, where each step corresponds to a mini-batch
update. As a test set, 100 block-placing tasks were employed.
At the end of the training,
both agents appear to level out at roughly the same scores. The DQN agent overestimates the Q-values in the first 1.7 million training steps. Although reducing the learning rate helped weakening this effect, it slowed down the training process. This is a known problem with Q-learning.

As for the model-based approach, it can quickly learn a meaningful model that can achieve good results with MCTS\@. This suggests that learning the transition function is an easier task than learning Q-values for this block-placing problem.
The main pitfall of the model-based approach lies in accumulating errors for rollouts that reach many steps ahead. In this particular block-placing task, MCTS trajectories are only 10 steps deep and Minecraft frames are rather structured, hence, the accumulating error problem is not that prominent.
The slight
increase in variance of the reward predictions of future steps is alleviated by
using a discount rate of 0.95 for the MCTS agent as compared to a discount rate of
0.99 used by DQN\@. This value was found empirically to provide the best results.
\begin{center}
\begin{tabular}{l c c c c c c c}
        \toprule
        &\multicolumn{3}{c}{Average Reward} & \phantom{.} & \multicolumn{3}{c}{Success Rate}\\
\cmidrule{2-4} \cmidrule{6-8}
Agent     & 1M steps & 2.5M steps & 5M steps & & 1M steps & 2.5M steps & 5M steps\\
\midrule
MCTS                   & \textbf{2.15} & \textbf{2.36} & 2.42 & & \textbf{0.64} & \textbf{0.70} & 0.72\\
MCTS, no 1-ahead reward & 0.68 & 0.84 & 0.87 & & 0.27 & 0.30 & 0.31\\
DQN                    & 0.05 & 2.22 & \textbf{2.59} & & 0.12 & 0.64 & \textbf{0.74}\\
\bottomrule
\end{tabular}
\captionof{table}{Results of Different Algorithms for the Block-Placing Task in Minecraft (cf.\ smoothed values in Fig.~\ref{fig:mctsdqn}).}\label{tab:mctssearch}
\end{center}

Table~\ref{tab:mctssearch} demonstrates the smoothed scores for both agents after 1, 2.5, and 5 million training steps. After 1 million
steps, the model-based approach can already solve a considerate amount of
block-placing tasks whereas the DQN agent has not learned a reasonable
policy yet. For the DQN agent to catch up with the MCTS agent, it needs 1.5 million additional training steps, what underlines the model-based approach's data efficiency. In the end, DQN beats the MCTS agent only by a small margin (74\% vs. 72\% success rate).
Additionally, the table includes the MCTS agent's scores if the one-step-ahead prediction of the reward is not employed. 
During the tree search, this agent chooses a random action of unexplored
future states instead of greedily choosing the action with maximum immediate
reward. For the block-placing task, using one-step-ahead predicted rewards
doubles the score across training steps, i.e., scores comparable with the DQN
agent become achievable with only 100 trajectories.

\section{Conclusion}
In this paper, we explored the idea of creating a model-based reinforcement learning agent that could perform competitively with model-free methods, DQN, in particular.
To implement such an agent, a synthesis of learning a transition model with a
deep neural network and MCTS was developed. Our tests on a block-placing task
in Minecraft show that learning a meaningful transition model requires
considerably less training data than learning Q-values of comparable scores
with DQN\@. As the MCTS agent uses a tree search for finding the best
action, it takes longer to perform one action in comparison with DQN\@.
Therefore, our approach is interesting for cases where obtaining training
samples from the environment is costly. The nature of the block-placing task
justifies the greedy choice of immediate rewards and, hence, application of the
one-step-ahead prediction significantly improves the score performance. The
transition model suffers from the limited information of the last four input
frames: if past information becomes unavailable (e.g., no longer visible in
the last four frames) the model makes incorrect predictions about the environment
leading to suboptimal actions. Further research in using a recurrent neural
network could help eliminating this issue. 

\printbibliography{}
\end{document}

%% file: mcfig.tex
\begin{figure}[t!]
	\centering
	\begin{tikzpicture}

\node[inner sep=0pt] at (3.45,-1.28) {\includegraphics[width=2.29cm]{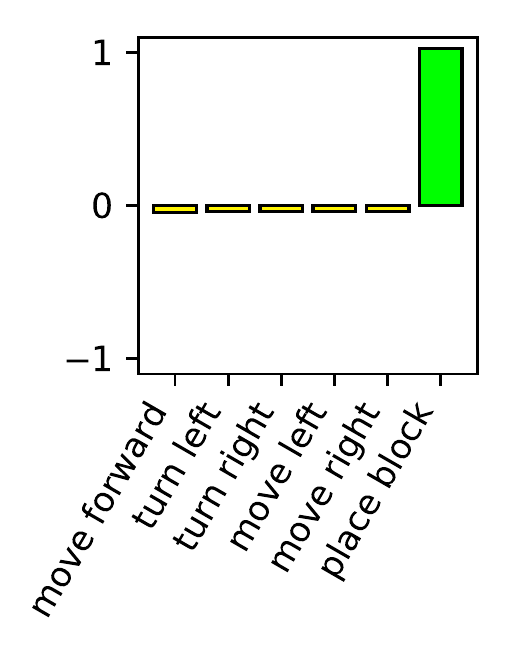}};
\node[inner sep=0pt] at (7.45,-1.28) {\includegraphics[width=2.29cm]{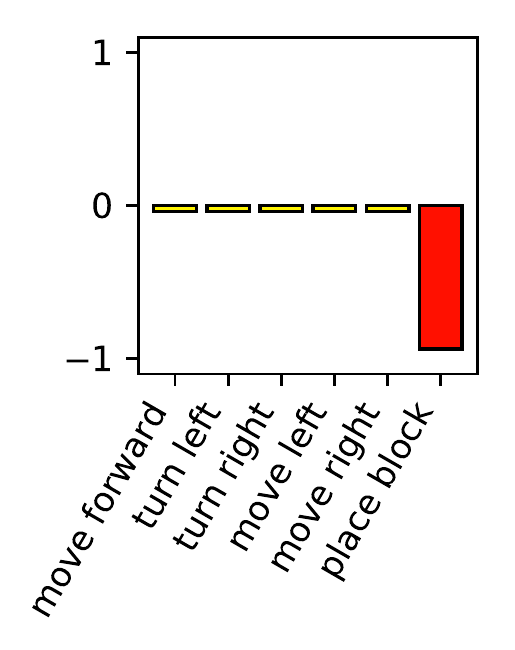}};

\fill [black, opacity=0.1] (2,1.1) rectangle (10.25,1.9);
\fill[black,opacity=0.1] (10.25,0.5) -- (10.25,2.5) -- (11,1.5);
    
\node[inner sep=0pt] at (0,1.5) {\includegraphics[width=4cm]{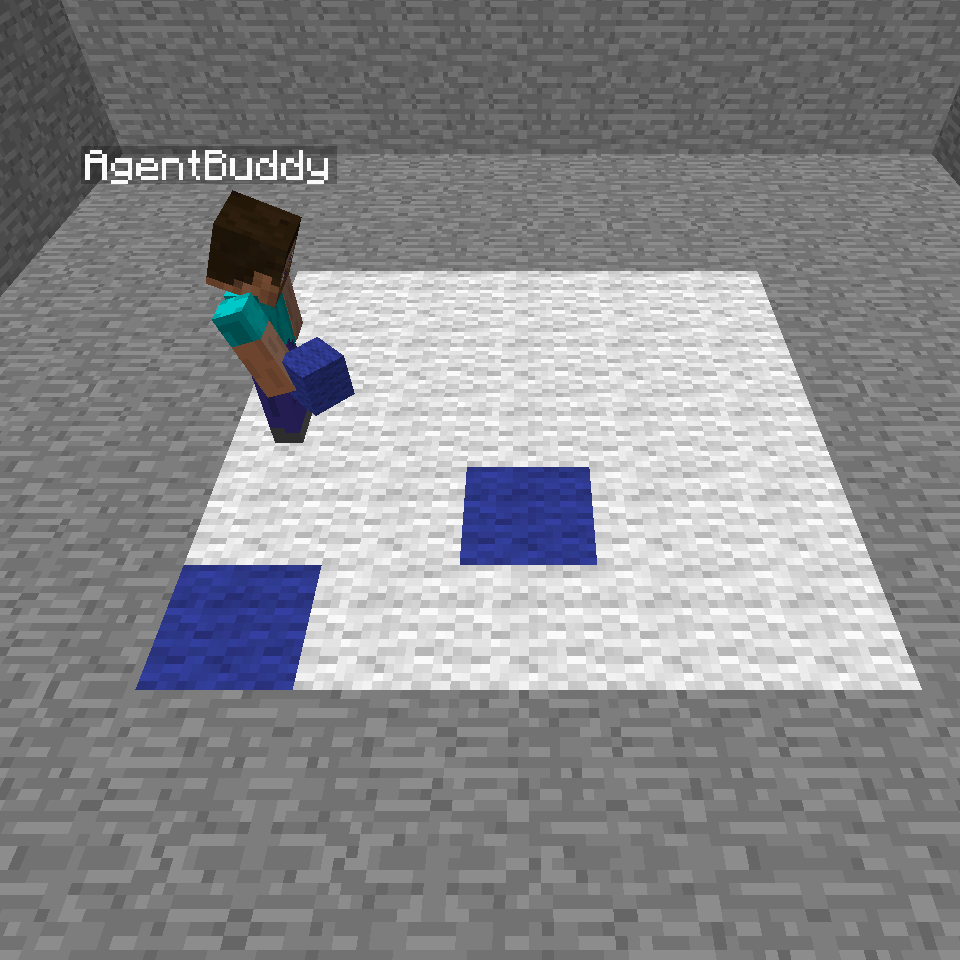}};
\node at (0,-1){\small Spectator View at Time Step \(t=3\)};

\node at (3.6,4.3){\large \(t=3\)};

\node[inner sep=0pt] at (5.25,3.75) {\includegraphics[width=1.5cm]{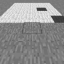}};
\draw (4.5, 3.0) -- (4.5, 4.5) -- (6.0, 4.5) -- (6.0, 3.0) -- (4.5, 3.0);
\node[inner sep=0pt] at (4.75,3.25) {\includegraphics[width=1.5cm]{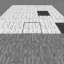}};
\draw (4.0, 2.5) -- (4.0, 4.0) -- (5.5, 4.0) -- (5.5, 2.5) -- (4.0, 2.5);
\node[inner sep=0pt] at (4.25,2.75) {\includegraphics[width=1.5cm]{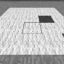}};
\draw (3.5, 2.0) -- (3.5, 3.5) -- (5.0, 3.5) -- (5.0, 2.0) -- (3.5, 2.0);
\node[inner sep=0pt] at (3.75,2.25) {\includegraphics[width=1.5cm]{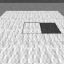}};
\draw (3.0, 1.5) -- (3.0, 3.0) -- (4.5, 3.0) -- (4.5, 1.5) -- (3.0, 1.5);

\draw (4.75,4.25) node{\small \contour{white}{\(s_0\)}};
\draw (4.25,3.75) node{\small \contour{white}{\(\widehat{s}_1\)}};
\draw (3.75,3.25) node{\small \contour{white}{\(\widehat{s}_2\)}};
\draw (3.25,2.75) node{\small \contour{white}{\(\widehat{s}_3\)}};

\node at (5.5,1.5){\small \begin{tabular}{c}\(a_3 = \)\\``turn right''\end{tabular}};
\draw[thick,->] (4.5,1.25) -- (4.5,0.25);
\node at (3.75,0.6){\small \begin{tabular}{c}DNN\\Model\end{tabular}};

\node[inner sep=0pt] (prediction) at (5.35,-0.75) {\includegraphics[width=1.5cm]{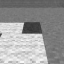}};
\draw (4.6, -1.5) -- (4.6, 0) -- (6.1, 0) -- (6.1, -1.5) -- (4.6, -1.5);
\node at (4.85,-0.25){\small \contour{white}{\(\widehat{s}_4\)}};
\node at (3.2,-0.25){\small \(\widehat{r}_5\)};

\node at (7.6,4.3){\large \(t=4\)};

\node[inner sep=0pt] at (9.25,3.75) {\includegraphics[width=1.5cm]{figures/pred_v2_0}};
\draw (8.5, 3.0) -- (8.5, 4.5) -- (10.0, 4.5) -- (10.0, 3.0) -- (8.5, 3.0);
\node[inner sep=0pt] at (8.75,3.25) {\includegraphics[width=1.5cm]{figures/pred_v2_1}};
\draw (8.0, 2.5) -- (8.0, 4.0) -- (9.5, 4.0) -- (9.5, 2.5) -- (8.0, 2.5);
\node[inner sep=0pt] at (8.25,2.75) {\includegraphics[width=1.5cm]{figures/pred_v2_2}};
\draw (7.5, 2.0) -- (7.5, 3.5) -- (9.0, 3.5) -- (9.0, 2.0) -- (7.5, 2.0);
\node[inner sep=0pt] (predinput) at (7.75,2.25) {\includegraphics[width=1.5cm]{figures/pred_v2_3}};
\draw (7.0, 1.5) -- (7.0, 3.0) -- (8.5, 3.0) -- (8.5, 1.5) -- (7.0, 1.5);

\draw (8.75,4.25) node{\small \contour{white}{\(\widehat{s}_1\)}};
\draw (8.25,3.75) node{\small \contour{white}{\(\widehat{s}_2\)}};
\draw (7.75,3.25) node{\small \contour{white}{\(\widehat{s}_3\)}};
\draw (7.25,2.75) node{\small \contour{white}{\(\widehat{s}_4\)}};

\draw[->,dashed,thick] (prediction.north east) -- (predinput.south west);


\node at (9.5,1.5){\small \begin{tabular}{c}\(a_4 = \)\\``place block''\end{tabular}};
\draw[thick,->] (8.5,1.25) -- (8.5,0.25);
\node at (7.75,0.6){\small \begin{tabular}{c}DNN\\Model\end{tabular}};

\node[inner sep=0pt] at (9.35,-0.75) {\includegraphics[width=1.5cm]{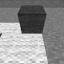}};
\draw (8.6, -1.5) -- (8.6, 0) -- (10.1, 0) -- (10.1, -1.5) -- (8.6, -1.5);
\node at (8.85,-0.25){\small \contour{white}{\(\widehat{s}_5\)}};
\node at (7.2,-0.25){\small \(\widehat{r}_6\)};

\node[inner sep=0pt] at (13,1.5) {\includegraphics[width=4cm]{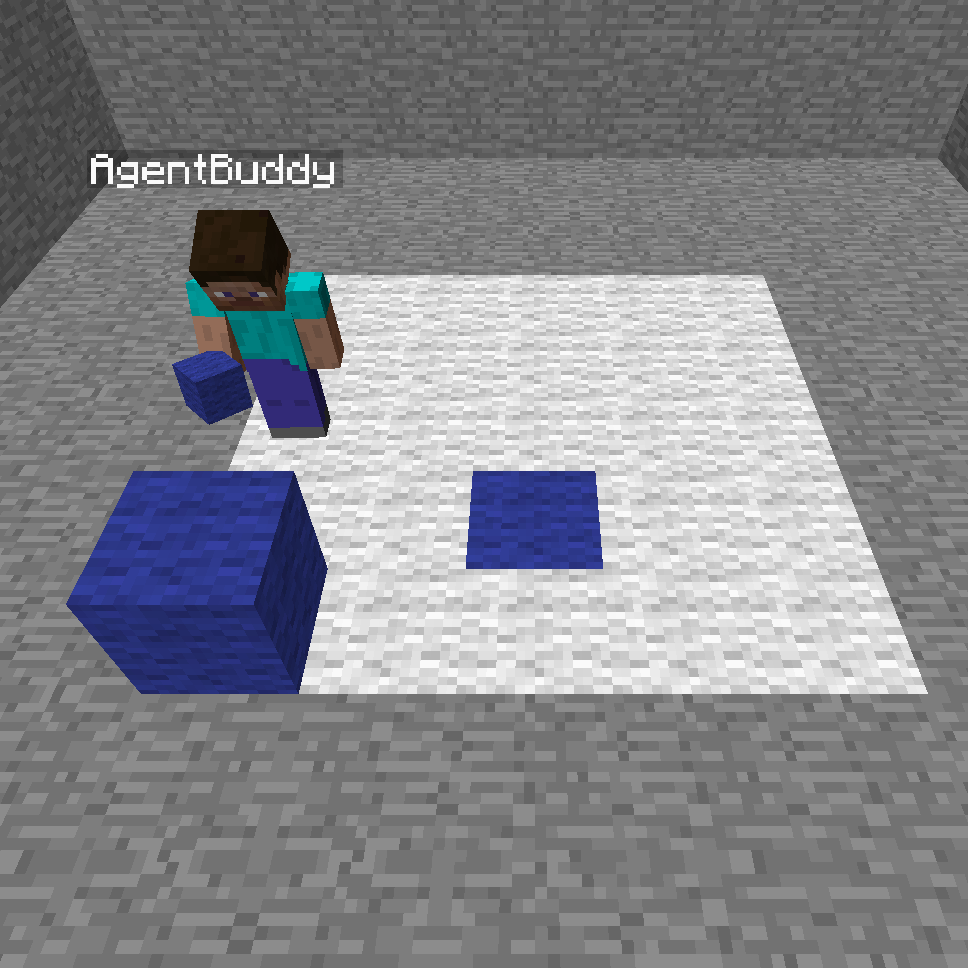}};
\node at (13,-1){\small Spectator View at Time Step \(t=5\)};

	\end{tikzpicture}
    \vspace{-2.5mm}
    \caption{Transition Model for the Block-Placing Task in Minecraft}\label{fig:minecraft}
\end{figure}

%% file: dnn.tex
\begin{figure}[t!]
	\centering
	\begin{tikzpicture}

        \draw (8, 2.5) -- (8.5, 2);
        \draw (7.75, 2.25) -- (7, 0.5);
        
        \draw (6.625, 1.375) -- (8.5, 2);
        \draw (5.5, 0.0) -- (7, 0.5);

        \draw[dashed] (7.5, 0.5) -- (8.5, 0);
        \draw[dashed] (9, 2) -- (9.875, 1.375);

        \draw (7.5, 0.5) -- (9.75, 1.75);
        \draw (9, 2) -- (10.5, 2.5);

\draw (7.5,1.7) node{\contour{white}{\scriptsize Affine}};
\draw (9.5,2) node{\contour{white}{\scriptsize Affine}};
\draw (11.25,2.25) node{\scriptsize Affine};
\draw (8.25,0.625) node{\contour{white}{\scriptsize Reshape}};

        \node at (4,-1.25){\small \begin{tabular}{c}Convolutional Layers\\\(32 \times 32 \times 32\)\hspace*{1.5cm}\\\hspace*{0.5cm}\(\rightarrow 64 \times 16 \times 16\)\hspace*{1cm}\\\hspace*{1cm}\(\rightarrow 128 \times 8 \times 8\)\hspace*{0.5cm}\\\hspace*{1.5cm}\(\rightarrow 256 \times 4 \times 4\)\end{tabular}};

        \node at (4.5,2.55){\scriptsize \begingroup\setlength{\fboxsep}{0pt}\colorbox{lightgray}{\begin{tabular}{c c c}\toprule[0.1ex] Filter & Stride & Pad.\\\midrule[0.1ex] \(7 \times 7\)& 2 & 3\\\(5 \times 5\) & 2 & 2\\\(5 \times 5\)& 2 & 2\\\(3 \times 3\) & 2 & 1\\\bottomrule[0.1ex] \end{tabular}}\endgroup};

        \node at (0.5,-0.72){\small \begin{tabular}{c}Pixel Input \(s_{t-3},\dots,s_t\)\\ \(4 \times 64 \times 64\) \end{tabular}};

    \node at (7,-0.72){\small \begin{tabular}{c}Fully Connected\\\(1 \times 4096\)\end{tabular}};
    \node at (10.25,-0.89){\small \begin{tabular}{c}Deconvolutional Layers\\(Same Dimensions\\in Reversed Order)\end{tabular}};
		
    \node at (7.75,3.025){\small \begin{tabular}{c}Action Input \(a_t\)\\ \(1 \times 6\) \end{tabular}};

    \node at (10.5,3.025){\small \begin{tabular}{c}Fully Connected\\\(1 \times 2048\)\end{tabular}};

    \node at (14,-0.72){\small \begin{tabular}{c}Pixel Prediction \(\widehat{s}_{t+1}\)\\\(1 \times 64 \times 64\)\end{tabular}};
    \node at (13.5,3.025){\small \begin{tabular}{c}Reward Prediction \(\widehat{r}_{t+2}\)\\\(1 \times 6\)\end{tabular}};

\node[inner sep=0pt] at (1.5,1.5)
    {\includegraphics[width=1.5cm]{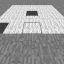}};
        \draw (0.75, 0.75) -- (0.75, 2.25) -- (2.25, 2.25) -- (2.25, 0.75) -- (0.75, 0.75);
\node[inner sep=0pt] at (1.25,1.25)
    {\includegraphics[width=1.5cm]{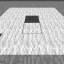}};
        \draw (0.5, 0.5) -- (0.5, 2.0) -- (2.0, 2.0) -- (2.0, 0.5) -- (0.5, 0.5);
\node[inner sep=0pt] at (1,1)
    {\includegraphics[width=1.5cm]{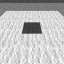}};
        \draw (0.25, 0.25) -- (0.25, 1.75) -- (1.75, 1.75) -- (1.75, 0.25) -- (0.25, 0.25);
\node[inner sep=0pt] at (0.75,0.75)
    {\includegraphics[width=1.5cm]{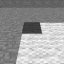}};
        \draw (0, 0) -- (0, 1.5) -- (1.5, 1.5) -- (1.5, 0) -- (0, 0);

\draw[fill=white] (3.25, 0.75) -- (3.25, 1.5) -- (4.0, 1.5) -- (4.0, 0.75) -- (3.25, 0.75);
\draw[fill=white] (3.1, 0.6) -- (3.1, 1.35) -- (3.85, 1.35) -- (3.85, 0.6) -- (3.1, 0.6);
\draw[fill=white] (2.95, 0.45) -- (2.95, 1.2) -- (3.7, 1.2) -- (3.7, 0.45) -- (2.95, 0.45);
\draw[fill=white] (2.8, 0.3) -- (2.8, 1.05) -- (3.55, 1.05) -- (3.55, 0.3) -- (2.8, 0.3);
\draw[fill=white] (2.65, 0.15) -- (2.65, 0.9) -- (3.4, 0.9) -- (3.4, 0.15) -- (2.65, 0.15);
\draw[fill=white] (2.5, 0.0) -- (2.5, 0.75) -- (3.25, 0.75) -- (3.25, 0.0) -- (2.5, 0.0);
		

		
\node at (4.9,0.8){\huge \(\dots\)};


\draw[fill=white] (6.375, 1.125) -- (6.375, 1.375) -- (6.625, 1.375) -- (6.625, 1.125) -- (6.375, 1.125);
\draw[fill=white] (6.3, 1.05) -- (6.3, 1.3) -- (6.55, 1.3) -- (6.55, 1.05) -- (6.3, 1.05);
\draw[fill=white] (6.225, 0.975) -- (6.225, 1.225) -- (6.475, 1.225) -- (6.475, 0.975) -- (6.225, 0.975);
\draw[fill=white] (6.15, 0.9) -- (6.15, 1.15) -- (6.4, 1.15) -- (6.4, 0.9) -- (6.15, 0.9);
\draw[fill=white] (6.075, 0.825) -- (6.075, 1.075) -- (6.325, 1.075) -- (6.325, 0.825) -- (6.075, 0.825);
\draw[fill=white] (6.0, 0.75) -- (6.0, 1.0) -- (6.25, 1.0) -- (6.25, 0.75) -- (6.0, 0.75);
\draw[fill=white] (5.925, 0.675) -- (5.925, 0.925) -- (6.175, 0.925) -- (6.175, 0.675) -- (5.925, 0.675);
\draw[fill=white] (5.85, 0.6) -- (5.85, 0.85) -- (6.1, 0.85) -- (6.1, 0.6) -- (5.85, 0.6);
\draw[fill=white] (5.775, 0.525) -- (5.775, 0.775) -- (6.025, 0.775) -- (6.025, 0.525) -- (5.775, 0.525);
\draw[fill=white] (5.7, 0.45) -- (5.7, 0.7) -- (5.95, 0.7) -- (5.95, 0.45) -- (5.7, 0.45);
\draw[fill=white] (5.625, 0.375) -- (5.625, 0.625) -- (5.875, 0.625) -- (5.875, 0.375) -- (5.625, 0.375);
\draw[fill=white] (5.55, 0.3) -- (5.55, 0.55) -- (5.8, 0.55) -- (5.8, 0.3) -- (5.55, 0.3);
\draw[fill=white] (5.475, 0.225) -- (5.475, 0.475) -- (5.725, 0.475) -- (5.725, 0.225) -- (5.475, 0.225);
\draw[fill=white] (5.4, 0.15) -- (5.4, 0.4) -- (5.65, 0.4) -- (5.65, 0.15) -- (5.4, 0.15);
\draw[fill=white] (5.325, 0.075) -- (5.325, 0.325) -- (5.575, 0.325) -- (5.575, 0.075) -- (5.325, 0.075);
\draw[fill=white] (5.25, 0.0) -- (5.25, 0.25) -- (5.5, 0.25) -- (5.5, 0.0) -- (5.25, 0.0);

		\draw[fill=black,draw=black,opacity=0.3] (7.25,2.25) -- (7.75,2.25) -- (8,2.5) -- (7.5,2.5) -- (7.25,2.25);

		\draw[fill=black,draw=black,opacity=0.3] (7,0.5) -- (7.5,0.5) -- (9,2) -- (8.5,2) -- (7,0.5);

\draw[fill=white] (9.625, 1.125) -- (9.625, 1.375) -- (9.875, 1.375) -- (9.875, 1.125) -- (9.625, 1.125);
\draw[fill=white] (9.55, 1.05) -- (9.55, 1.3) -- (9.8, 1.3) -- (9.8, 1.05) -- (9.55, 1.05);
\draw[fill=white] (9.475, 0.975) -- (9.475, 1.225) -- (9.725, 1.225) -- (9.725, 0.975) -- (9.475, 0.975);
\draw[fill=white] (9.4, 0.9) -- (9.4, 1.15) -- (9.65, 1.15) -- (9.65, 0.9) -- (9.4, 0.9);
\draw[fill=white] (9.325, 0.825) -- (9.325, 1.075) -- (9.575, 1.075) -- (9.575, 0.825) -- (9.325, 0.825);
\draw[fill=white] (9.25, 0.75) -- (9.25, 1.0) -- (9.5, 1.0) -- (9.5, 0.75) -- (9.25, 0.75);
\draw[fill=white] (9.175, 0.675) -- (9.175, 0.925) -- (9.425, 0.925) -- (9.425, 0.675) -- (9.175, 0.675);
\draw[fill=white] (9.1, 0.6) -- (9.1, 0.85) -- (9.35, 0.85) -- (9.35, 0.6) -- (9.1, 0.6);
\draw[fill=white] (9.025, 0.525) -- (9.025, 0.775) -- (9.275, 0.775) -- (9.275, 0.525) -- (9.025, 0.525);
\draw[fill=white] (8.95, 0.45) -- (8.95, 0.7) -- (9.2, 0.7) -- (9.2, 0.45) -- (8.95, 0.45);
\draw[fill=white] (8.875, 0.375) -- (8.875, 0.625) -- (9.125, 0.625) -- (9.125, 0.375) -- (8.875, 0.375);
\draw[fill=white] (8.8, 0.3) -- (8.8, 0.55) -- (9.05, 0.55) -- (9.05, 0.3) -- (8.8, 0.3);
\draw[fill=white] (8.725, 0.225) -- (8.725, 0.475) -- (8.975, 0.475) -- (8.975, 0.225) -- (8.725, 0.225);
\draw[fill=white] (8.65, 0.15) -- (8.65, 0.4) -- (8.9, 0.4) -- (8.9, 0.15) -- (8.65, 0.15);
\draw[fill=white] (8.575, 0.075) -- (8.575, 0.325) -- (8.825, 0.325) -- (8.825, 0.075) -- (8.575, 0.075);
\draw[fill=white] (8.5, 0.0) -- (8.5, 0.25) -- (8.75, 0.25) -- (8.75, 0.0) -- (8.5, 0.0);

\node at (10.4,0.8){\huge \(\dots\)};

\draw[fill=white] (11.75, 0.75) -- (11.75, 1.5) -- (12.5, 1.5) -- (12.5, 0.75) -- (11.75, 0.75);
\draw[fill=white] (11.6, 0.6) -- (11.6, 1.35) -- (12.35, 1.35) -- (12.35, 0.6) -- (11.6, 0.6);
\draw[fill=white] (11.45, 0.45) -- (11.45, 1.2) -- (12.2, 1.2) -- (12.2, 0.45) -- (11.45, 0.45);
\draw[fill=white] (11.3, 0.3) -- (11.3, 1.05) -- (12.05, 1.05) -- (12.05, 0.3) -- (11.3, 0.3);
\draw[fill=white] (11.15, 0.15) -- (11.15, 0.9) -- (11.9, 0.9) -- (11.9, 0.15) -- (11.15, 0.15);
\draw[fill=white] (11.0, 0.0) -- (11.0, 0.75) -- (11.75, 0.75) -- (11.75, 0.0) -- (11.0, 0.0);

\node[inner sep=0pt] at (14,0.75)
    {\includegraphics[width=1.5cm]{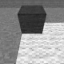}};
\draw (13.25, 0) -- (13.25, 1.5) -- (14.75, 1.5) -- (14.75, 0) -- (13.25, 0);

\draw (0.6, 0.15) -- (0.6, 0.45) -- (0.9, 0.45) -- (0.9, 0.15) -- (0.6, 0.15);
\draw (0.9, 0.45) -- (2.875, 0.15);
\draw (0.6, 0.45) -- (2.875, 0.15);
\draw (0.6, 0.15) -- (2.875, 0.15);
\draw (0.9, 0.15) -- (2.875, 0.15);

\draw (11.3, 0.45) -- (11.3, 0.6) -- (11.45, 0.6) -- (11.45, 0.45) -- (11.3, 0.45);
\draw (11.3, 0.6) -- (8.625, 0.175);
\draw (11.45, 0.6) -- (8.625, 0.175);
\draw (11.3, 0.45) -- (8.625, 0.175);
\draw (11.45, 0.45) -- (8.625, 0.175);
\fill [white] (9.6,0.15) -- (9.85,0.4) -- (10.5,0.5) -- (10.25,0.25);
\draw[loosely dashed] (11.3, 0.6) -- (8.625, 0.175);
\draw[loosely dashed] (11.45, 0.45) -- (8.625, 0.175);

\draw (2.8, 0.45) -- (2.8, 0.6) -- (2.95, 0.6) -- (2.95, 0.45) -- (2.8, 0.45);
\draw (2.8, 0.6) -- (5.375, 0.175);
\draw (2.95, 0.6) -- (5.375, 0.175);
\draw (2.8, 0.45) -- (5.375, 0.175);
\draw (2.95, 0.45) -- (5.375, 0.175);
\fill [white] (4.25,0.45) -- (4,0.2) -- (4.6,0.1) -- (4.85,0.35);
\draw[loosely dashed] (5.375, 0.175) -- (2.8, 0.6);
\draw[loosely dashed] (2.95, 0.45) -- (5.375, 0.175);

\draw (13.85, 0.15) -- (13.85, 0.45) -- (14.15, 0.45) -- (14.15, 0.15) -- (13.85, 0.15);
\draw (13.85, 0.15) -- (11.375, 0.15);
\draw (13.85, 0.45) -- (11.375, 0.15);
\draw (14.15, 0.45) -- (11.375, 0.15);
\draw (13.85, 0.15) -- (11.375, 0.15);

\draw[fill=black,draw=black,opacity=0.3] (9.75,1.75) -- (10.25,1.75) -- (11,2.5) -- (10.5,2.5) -- (9.75,1.75);

        \draw (10.25, 1.75) -- (13, 2.25);
        \draw (11, 2.5) -- (13.25, 2.5);

\draw[fill=black,draw=black,opacity=0.3] (13,2.25) -- (13.5,2.25) -- (13.75,2.5) -- (13.25,2.5) -- (13,2.25);

	\end{tikzpicture}
    \vspace{-2.5mm}
    \caption{Architecture of the Transition Model}\label{fig:dnnarch}
\end{figure}